\documentclass{article}

\PassOptionsToPackage{numbers,compress}{natbib}
\usepackage[final,main]{neurips_2026}

\usepackage[utf8]{inputenc}
\usepackage[T1]{fontenc}
\usepackage{hyperref}
\usepackage{url}
\usepackage{booktabs}
\usepackage{amsfonts}
\usepackage{nicefrac}
\usepackage{microtype}
\usepackage{xcolor}

\usepackage{graphicx}
\usepackage{subcaption}
\usepackage{amsmath}
\usepackage{amssymb}
\usepackage{mathtools}
\usepackage{amsthm}
\usepackage[capitalize,noabbrev]{cleveref}

\usepackage{multirow}
\usepackage{multicol}
\usepackage[export]{adjustbox}
\usepackage[table]{colortbl}
\usepackage{wrapfig}
\usepackage{enumitem}
\usepackage{tcolorbox}
\usepackage{algorithm}
\usepackage{algorithmic}

\hypersetup{hidelinks}

\theoremstyle{plain}

\theoremstyle{definition}

\theoremstyle{remark}

\title{3A-VLA: Abstraction-Aligned Action Learning for
Vision-Language Agents in 3D Game Worlds}

\author{%
  \textbf{Zheyuan Zhou}$^{1,2}$\thanks{Equal contribution. \qquad
  $\dagger$ Corresponding author.} \quad
  \textbf{Liang Du}$^{1,*}$ \quad
  \textbf{Zixun Sun}$^{1,2,*,\dagger}$ \\[3pt]
  \textbf{Xiaoyu Zhou}$^{1}$ \quad
  \textbf{Ruimin Ye}$^{1}$ \quad
  \textbf{Qihao Chen}$^{1}$ \quad
  \textbf{Yinda Chen}$^{1}$ \quad
  \textbf{Lemiao Qiu}$^{2}$ \\[7pt]
  $^{1}$IEG, Tencent
  \qquad
  $^{2}$Zhejiang University
}

\begin{document}

\maketitle

\vspace{-0.22in}
{\centering
\small
\url{https://3a-vla.github.io/}\par
}
\vspace{0.05in}

\begin{abstract}
Despite advances in vision-language-action (VLA) models, they remain limited in highly dynamic game settings such as 3D open worlds and competitive player versus player (PvP) settings, where agents must efficiently extract sparse, actionable signals from dense visual input while integrating multimodal cues to track off-screen high-value targets in real time.
To address this, we introduce \textbf{3A-VLA} (\textbf{A}bstraction-\textbf{A}ligned \textbf{A}ction VLA), a framework that grounds action learning in explicit intention and environment abstractions rather than superficial pattern matching.
We introduce dual task-agnostic abstractions: the intention abstraction (IA), which condenses verbose instructions and reasoning into explicit semantic primitives, and the environment semantics abstraction (ESA), which structures dense visual streams into a spatial--functional affordance representation to guide grounded actions.
We further propose an abstraction alignment reweighting (AAR) module that adaptively reweights the action imitation loss. A continuous intention--environment alignment signal is used to emphasize reliable action supervision when abstractions agree and reduce the influence of ambiguous demonstrations when they diverge, thereby learning a policy that balances fine-grained control and high-level reasoning without the need for manual rules.
Extensive experiments show that 3A-VLA performs competitively among
the evaluated baselines in both open-world (Minecraft) and competitive PvP (Game for Peace) settings. 
It also demonstrates zero-shot combat transfer from Game for Peace to high-fidelity games, including Valorant, CS2, GTA V, Elden Ring, and Mount \& Blade.
\end{abstract}
\begin{figure}[t]
\centering
\includegraphics[width=1\linewidth]{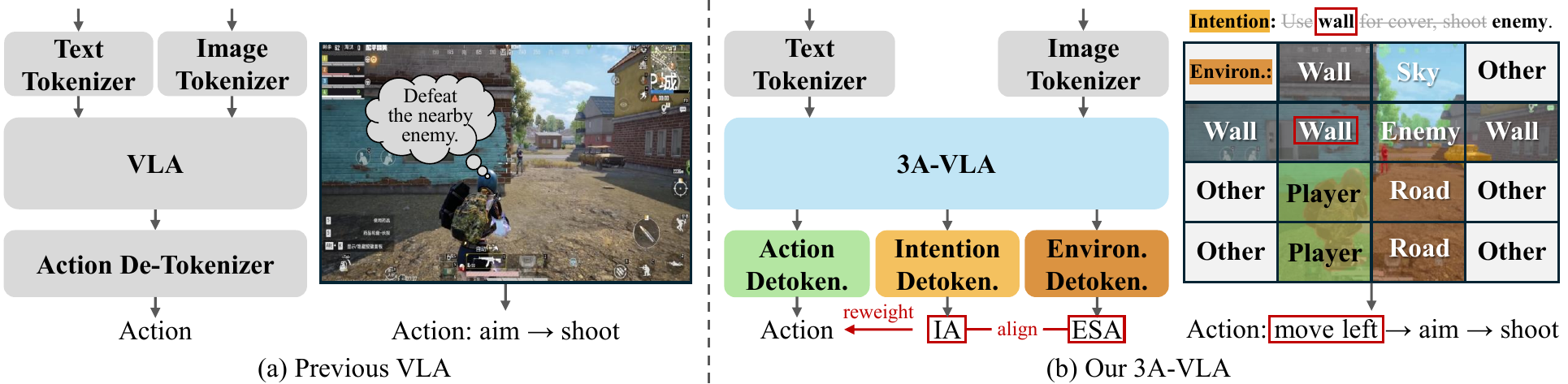}
\caption{
Unlike previous VLA models which mapped low-level inputs directly to actions, 3A-VLA aligns action learning with intention abstraction (IA) and environment semantics abstraction (ESA) through multimodal detokenizers.
As shown in a PvP scenario, 3A-VLA grounds instructions and visual scenes into high-level, interpretable concepts such as ``wall'' and ``enemy''.
%
\textbf{(i)} When trained on action sequences such as ``move left, aim, shoot,'' the model learns the semantic relationship between ``wall'' and ``move left,'' recognizing the object's functional role as tactical cover rather than merely its presence.
\textbf{(ii)} When tasked with locating off-screen enemies, the model uses the structured affordance representation from the ESA to reason about likely ambush points, such as inferring that a ``wall'' may serve as high-probability cover from which an opponent could emerge, thereby enabling proactive searches.
In off-screen scenarios with weak IA-ESA correlation, semantic reasoning outweighs action prediction; 3A-VLA adapts by reweighting action learning to prioritize semantic grounding.
}
\label{fig:teaser}
\end{figure}

\section{Introduction}
Recent vision-language-action (VLA) models have enabled embodied agents to interpret multimodal instructions and execute complex tasks within immersive digital environments~\cite{vpt, wang2023voyager, gametars}. 
These methods have demonstrated strong performance in structured, relatively static embodied settings
~\cite{brohan2023rt1, zitkovich2023rt2, kim2024openvla} and structured yet varied interactions in PvE game environments~\cite{chen2025combatvla,tan2025lumine}.
However, in dynamic game environments defined by completely spontaneous, real-time interactions, existing models still face a fundamental challenge: perceptual overload~\cite{brokenvideos}. 
To navigate an open world with emergent events (e.g., suddenly appearing vehicles) or a competitive player versus player (PvP) match with sudden threats (e.g., abruptly emerging opponents), the model must continuously process complex sensory data while parsing instructions that are often verbose.
Existing approaches lack the capacity to efficiently exploit the most essential signals for decision-making from this large amount of data with these signals remaining implicitly embedded and difficult to access without explicit selection mechanisms~\cite{coarse}.
This deficiency is rooted in a profound modality misalignment between continuous visual features and discrete language tokens~\cite{clip, zhai2023sigmoid, mindthegap, zitkovich2023rt2}.
Owing to this substantial semantic gap, existing models struggle to align high-level instructions with low-level observations. 
This forces agents to rely on superficial pattern matching, often referred to as ``instruction overfitting'' or ``causal confusion'', rather than achieving true cross-modal understanding. 
Ultimately, this results in fragile policies that fail to generalize, leading to unreliable decision-making in complex, open-world environments~\cite{langosco2022goal,fan2022minedojo}.
Minecraft and Game for Peace show roughly $2$--$3{\times}$ the intra-action visual diversity of LIBERO, measured by averaging pairwise cosine distances between CLIP features within action bins and weighting by bin frequency.

To bridge this gap, we draw inspiration from the cognitive and neural sciences, which reveal that the human brain does not process all sensory data equally~\cite{gwt}.
Instead, as shown in Figure~\ref{fig:framework}, it relies on a conscious bottleneck, that filters vast streams of irrelevant information and abstracts high-level semantics to extract only actionable primitives.
This process allows the brain to ignore the noise of detailed textures and wordy sentences and instead focus on sparse, abstract representations of reality.
We posit that for embodied agents to move beyond the following of simple instructions to truly understand intent, they must incorporate a similar inductive bias: learning to actively abstract the hidden, intention-derived information from redundant multimodal data streams.
Guided by this principle, we introduce \textbf{3A-VLA} (\textbf{A}bstraction-\textbf{A}ligned \textbf{A}ction VLA).
Unlike standard VLA models that passively map sensory inputs to actions, 3A-VLA trains the model to align action learning with task-relevant abstractions from noisy multimodal data streams.
We instantiate this framework via a unified causal Transformer~\cite{qwen2vl} that acts as a cognition-inspired bottleneck, synthesizing two complementary abstraction pathways.
First, through the intention abstraction (IA), the model aligns its internal representations with compact intention primitives
(e.g., the concept of a ``wall'' in the text in Figure~\ref{fig:teaser}).
This process involves extracting the latent strategic intent from verbose instructions, encouraging the agent to internalize a high-level task structure rather than relying on superficial phrase matching.
Simultaneously, to mitigate visual clutter, we employ the environment semantics abstraction (ESA).
By projecting visual streams into a sparse topological representation of task-critical affordances
(e.g., the entity ``wall'' in the semantic map in Figure~\ref{fig:teaser}),
our method prioritizes the functional structure of the environment over irrelevant textural details.
Both pathways are supervised by zero-shot off-the-shelf components (an open-source VLM for IA keywords and a pretrained segmenter for the ESA), with no retrieval, domain-specific fine-tuning, human annotation, or more powerful annotator model needed.
As a benefit of the abstraction procedure, the model's internal attention naturally concentrates on task-relevant regions, which enables a parameter-free token-pruning strategy that trims visual tokens at inference with negligible performance loss.

We further introduce the abstraction alignment loss reweighting (AAR) module, enabling an adaptive policy that ties action supervision to high-level semantic alignment. As shown in Figure~\ref{fig:teaser}, this governs situational dynamics ranging from in-view execution to off-screen search.
Specifically, AAR reweights the action imitation loss using a continuous alignment signal derived from the intention and environment abstractions. Rather than switching modes, the model emphasizes action supervision when alignment is strong (e.g., target in view), while reducing the influence of ambiguous demonstrations as alignment weakens (e.g., target absent, the scene is task-agnostic). This soft reweighting mechanism allows the policy to remain guided by semantic abstractions without the need for manual intervention.
Our contributions are summarized as follows:
\begin{itemize}
\item We propose \textbf{3A-VLA}, an \textbf{A}bstraction-\textbf{A}ligned \textbf{A}ction VLA framework that grounds decision-making in the alignment between the agent intention and the environment semantics, thereby increasing robustness over pattern-matching baselines in dynamic gaming environments. 
\item We propose a dual-channel abstraction to achieve semantic grounding. The intention abstraction (IA) compresses verbose instructions into actionable semantic primitives, whereas the environment semantics abstraction (ESA) aligns perceptual representations with the functional topology of the environment by encoding spatial affordances. Together, they form a coherent abstract state that enables robust situational reasoning.
\item We propose the abstraction alignment reweighting (AAR) module, which adaptively weights action imitation via a continuous intention--environment alignment signal, allowing the policy to emphasize reliable demonstrations while down-weighting ambiguous ones.
\item Extensive experiments demonstrate the state-of-the-art performance of 3A-VLA on open-world (Minecraft) and competitive PvP (Game for Peace) benchmarks, as well as its zero-shot combat transfer (trained on Game for Peace) to Valorant, CS2, GTA V, Elden Ring, Mount \& Blade, and Minecraft.
\end{itemize}

\section{Related work}

\subsection{Vision-Language-Action Models}
Vision-Language-Action (VLA) models ground visual and linguistic inputs into actions.
RT-1~\cite{brohan2023rt1} and RT-2~\cite{zitkovich2023rt2} established large-scale vision-language-action learning, while OpenVLA~\cite{kim2024openvla} and Octo~\cite{team2024octo} further improved generalization through scalable transformer architectures.
Diffusion Policy~\cite{chi2023diffusionpolicy} and the $\pi$-series~\cite{black2024pi0,intelligence2025pi} introduced generative action modeling for precise control.
Despite these advances, existing approaches often directly fuse multimodal inputs without explicitly separating task-relevant intention from environmental information, making them vulnerable to irrelevant details.
Embodied Chain-of-Thought (ECoT)~\cite{zawalski2024robotic} introduces intermediate reasoning but primarily operates at the token level.
In contrast, we introduce structured intention and environment abstractions to guide action learning.

\subsection{Information Selection and Reweighting}
Reducing redundant information has been explored at both inference and training stages.
FastV~\cite{chen2024fastv} prunes visual tokens based on attention scores, while ToMe~\cite{bolya2022token} merges redundant tokens to accelerate inference.
Meanwhile, L2RW~\cite{ren2018reweight}, Meta-Weight-Net~\cite{shu2019metaweight}, learned data manipulation~\cite{hu2019manipulation}, and Co-teaching~\cite{han2018coteaching} improve training through adaptive sample weighting or selection.
Our approach integrates task-aware abstractions with connectivity-based token selection and relation-conditioned action-loss reweighting, emphasizing task-relevant information for efficient and robust action learning.

\subsection{Embodied Agents in Digital Environments}
Digital games provide complex and reproducible environments for embodied agents.
Existing approaches broadly fall into API-based and pixel-based paradigms.
API-based agents, such as Voyager~\cite{wang2023voyager} and Ghost in the Minecraft~\cite{zhu2023ghost}, leverage LLMs for high-level planning but often rely on privileged interfaces.
Pixel-based agents, including VPT~\cite{vpt} and STEVE-1~\cite{lifshitz2023steve}, learn visuomotor control from raw observations but remain limited in semantic planning and visual robustness.
CombatVLA~\cite{chen2025combatvla} extends VLA methods to 3D combat, while Cradle~\cite{tan2025cradle}, JARVIS-VLA~\cite{li2025jarvis}, and OpenHA~\cite{wang2025openha} explore more general agentic frameworks.
However, real-time control and robustness remain challenging.
Our 3A-VLA bridges these paradigms by integrating high-level semantic abstractions with pixel-based control in a unified architecture.

\section{Methodology}
\label{sec:method}

\begin{figure*}[t]
\centering
\includegraphics[width=1.0\linewidth]{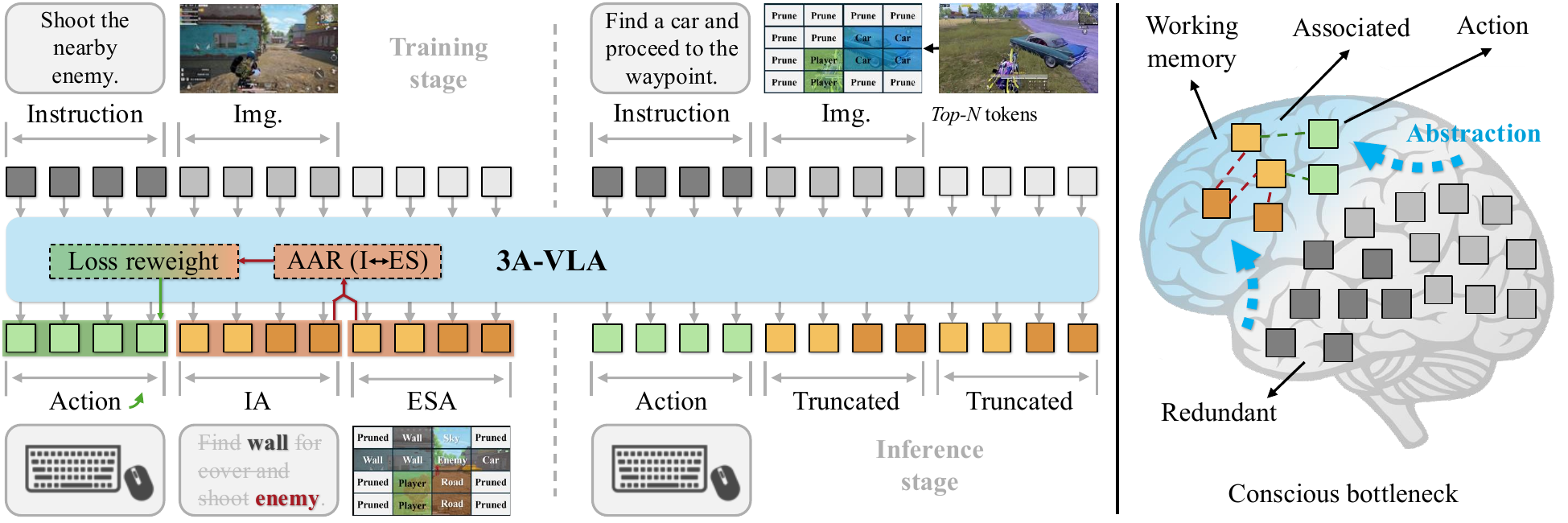}
\caption{\textbf{Overview.} 3A-VLA learns abstraction-aligned actions by coupling intention abstraction (IA) and environment semantics abstraction (ESA) with abstraction alignment reweighting (AAR). During training, IA and ESA expose sparse task primitives from language and vision, while AAR reweights action supervision according to intention--environment alignment. At inference, the induced sparsity enables low-latency control by keeping only top-$N$ task-critical visual tokens.}
\label{fig:framework}
\end{figure*}

\subsection{Overview}
\label{sec:overview}

As illustrated in Figure~\ref{fig:framework}, the agent receives an observation $\mathbf{x}_v$ and a language instruction $\mathbf{x}_l$; let $\mathbf{X}=(\mathbf{x}_v,\mathbf{x}_l)$, and let $\mathbf{A}$ denote the expert action. 
In dynamic environments, direct imitation learning strategies face two practical difficulties: (i) observations are dense and often dominated by task-irrelevant details such as texture, background motion. 
(ii) short-horizon action labels are often ambiguous. In particular, the same action may be demonstrated under different intents and scene contexts, while similar observations may require different actions depending on the task.
As an abstraction-aligned action learner, 3A-VLA augments a standard autoregressive VLA with two compact abstraction targets: an intention target $\mathbf{y}_{\mathrm{ia}}$ and an environment semantic target $\mathbf{y}_{\mathrm{esa}}$.
Specifically, $\mathbf{y}_{\mathrm{ia}}\in\mathcal{V}_{\mathrm{ia}}^{L_{\mathrm{ia}}}$ is a short sequence of intent keywords, and $\mathbf{y}_{\mathrm{esa}}$ is a flattened coarse semantic grid over tactical categories.
The IA summarizes the agent’s goal and strategy, whereas the ESA summarizes which task-relevant information is present in the visual scene.
Both targets serve solely as auxiliary training signals and are omitted at inference.
All the components are implemented with a single autoregressive causal Transformer $\theta$ initialized from Qwen2-VL~\cite{qwen2vl}.
The same backbone is used to predict expert actions with IA/ESA targets during training. 
During inference, the policy follows the standard VLA deployment pipeline and uses only the actions from the predictions.

\subsection{Intention abstraction (IA)}
\label{sec:ia}

Linguistic instructions in open-ended worlds are verbose and unstructured (e.g., ``\emph{Find a way to quickly enter the safe zone to escape the hazard zone}''), yet the policy-relevant intent is sparse and discrete (e.g., \texttt{[waypoint,\ safezone]}). 
The IA provides a compact intent summary as $\mathbf{y}_{\mathrm{ia}}\in\mathcal{V}_{\mathrm{ia}}^{L_{\mathrm{ia}}}$.

\paragraph{Zero-shot intent generation.}
Standard trajectory datasets do not provide intent labels. We therefore synthesize IA targets offline using a single off-the-shelf open-source VLM, Qwen2-VL-7B~\cite{qwen2vl}. 
In contrast to prior works~\cite{chen2025combatvla, tan2025lumine} that rely on verbose chain-of-thought (CoT) rationales for supervision, we leverage the VLM's structured output from a single inference pass: while the model generates both reasoning steps and conclusions, we retain only the final decisive keyword sequence for supervision. 

$\mathbf{y}_{\mathrm{ia}}$ encourages the shared representation to preserve task intent in a compact and stable form. 
Unlike free-form rationales and chain-of-thought annotations, the keyword supervision reduces redundancy and exposes the policy to reusable task primitives. 
Consequently, this compact semantic representation aligns naturally with ESA, enabling a unified latent space where linguistic intent and visual scene structure are coupled.

\subsection{Environment semantics abstraction (ESA)}
\label{sec:esa}

Pixel streams in embodied control are massively redundant; the important information for many short-horizon decisions is the topology of the task-relevant affordances over the per-pixel RGB values. 
The ESA provides a compact scene-level semantic target $\mathbf{y}_{\mathrm{esa}}$ that summarizes the functional structure of the current observation.

\paragraph{Latent semantic grid.}
We compress the image into an $H_g\!\times\!W_g$ semantic grid $\mathbf{G}_{\mathrm{sem}}$.
Ground-truth grids are obtained by running an off-the-shelf segmenter (SegFormer-B2~\cite{xie2021segformer}) on $\mathbf{x}_v$ and projecting the dense mask into five coarse categories: \texttt{person}, \texttt{vehicle}, \texttt{cover}, \texttt{item}, and \texttt{other}. 
No domain-specific fine-tuning is performed. 
Within each patch $\Omega_{u,v}$, we keep the top-$K$ classes under a tactical priority
$\rho(\texttt{person})>\rho(\texttt{vehicle})>\rho(\texttt{cover})>\rho(\texttt{item})>\rho(\texttt{other})$:
\begin{equation}
\mathbf{g}_{u,v} \;=\; \operatorname{Top}_K\bigl(\{c\mid c\in\Omega_{u,v}\},\,\rho\bigr).
\end{equation}
This priority-aware pooling strategy ensures that small but important foreground entities, such as distant people, are retained even when they occupy only a small fraction of a grid cell. 
We flatten $\mathbf{G}_{\mathrm{sem}}$ into an ordered token sequence $\mathbf{y}_{\mathrm{esa}}$ via a raster scan, with the $K$ class tokens inside each cell emitted in the order of their priority. The ordered-tuple convention provides a consistent tokenization for each grid cell and preserves the priority among cooccurring tactical categories.
$\mathbf{y}_{\mathrm{esa}}$ encourages the shared representation to focus on the functional scene structure rather than the raw texture information. 
The coarse grid format is intentionally sparse and task oriented. As shown in Table~\ref{tab:ablation}, denser alternatives such as action-conditioned image prediction and global semantic segmentation are less effective.

\subsection{Abstraction alignment reweighting (AAR)}
\label{sec:aar}

The IA and ESA describe complementary abstractions of each demonstration: IA specifies the intended objective, while ESA captures the relevant environmental affordances. 
The AAR reweights the action imitation loss according to the semantic alignment between these two abstractions.

Let $\mathbf{H}_i\in\mathbb{R}^{T_i\times d}$ be the final-layer hidden states for sample $i$. 
For $a\in\{\mathrm{ia},\mathrm{esa}\}$, let $\mathcal{T}^{a}_i$ denote the token span of the IA or ESA. 
We obtain span-level representations by mean pooling:
\begin{equation}
\mathbf{h}^{a}_i
=
\frac{1}{|\mathcal{T}^{a}_i|}
\sum_{t\in\mathcal{T}^{a}_i}
\mathbf{H}_{i,t},
\quad
a\in\{\mathrm{ia},\mathrm{esa}\},
\end{equation}
where both $\mathbf{h}^{\mathrm{ia}}_i$ and $\mathbf{h}^{\mathrm{esa}}_i$ lie in $\mathbb{R}^{d}$.
We construct the IA--ESA relation feature as
\begin{equation}
\mathbf{u}^{a}_i
=
\frac{
\mathcal{D}(\mathbf{h}^{a}_i)
}{
\left\|
\mathcal{D}(\mathbf{h}^{a}_i)
\right\|_2+\epsilon
},
\quad
a\in\{\mathrm{ia},\mathrm{esa}\},
\qquad
\mathbf{r}_i
=
\mathbf{u}^{\mathrm{ia}}_i
\odot
\mathbf{u}^{\mathrm{esa}}_i .
\end{equation}
The stop gradient operator $\mathcal{D}$ prevents gradients in the reweighting branch from directly updating the hidden states.
$\odot$ is the Hadamard product.
The element-wise product preserves dimension-wise IA--ESA interactions for the scorer.

An MLP-based scorer $g_{\psi}$ is employed to estimate the importance of each relation feature. 
Specifically, $g_{\psi}$ first projects the relation feature $\mathbf{r}_i \in \mathbb{R}^{d}$ into a lower-dimensional bottleneck space, and then processes it using two residual MLP layers with layer normalization. 
A linear projection head maps the encoded relation feature to a scalar score. 
The resulting scores are normalized across the batch via a softmax operation to yield batch-relative action weights:
\begin{equation}
q_i = g_{\psi}(\mathbf{r}_i),
\qquad
p_i = \frac{\exp(q_i)}{\sum_{j=1}^{B}\exp(q_j)},
\qquad
w_i = B\, p_i,
\end{equation}
where the scaling factor $B$ guarantees that $\frac{1}{B}\sum_{i=1}^{B} w_i = 1$, thereby preserving the overall magnitude of the action supervision signal.
To avoid degenerate concentration on a few samples, we regularize the normalized entropy $\bar{H}$ of the batch-relative distribution:
\begin{equation}
\mathcal{L}
=
\frac{1}{B}
\sum_{i=1}^{B}
\left[
w_i\mathcal{L}^{(i)}_{\mathrm{act}}
+
\mathcal{L}^{(i)}_{\mathrm{abs}}
\right]
-
\lambda_{\bar{H}}\bar{H}(\mathbf{p}), 
\qquad
\bar{H}(\mathbf{p})
=
\frac{
-\sum_{i=1}^{B}p_i\log(p_i+\epsilon)
}{
\log B
},
\end{equation}
where $\mathcal{L}^{(i)}_{\mathrm{act}}$ is the action imitation loss, and
$\mathcal{L}^{(i)}_{\mathrm{abs}}$ consists of the token-level IA and ESA losses. 
We fix $\lambda_{\bar{H}}=0.01$ for all experiments without task-specific tuning. 

\section{Experiments}
\label{sec:exp}

\subsection{Experimental setup}

\paragraph{Benchmarks.}
For \textbf{Minecraft}, we follow OpenHA~\cite{wang2025openha} and evaluate on MCU~\cite{zheng2025mcu}, comprising 800+ embodied, combat, and GUI tasks. We report Mini (10 representative tasks per domain) and All (800+ tasks) protocols in out-of-distribution environments with novel seeds.
\textbf{Game for Peace} covers six atomic tasks at intermediate difficulty (gold/silver tiers): precision parachuting, resource scavenging, combat engagement, teammate revival, vehicle acquisition, and strategic rotation.
For \textbf{cross-game combat transfer}, GP-combat-trained models are evaluated without target-game fine-tuning in Valorant, CS2, GTA V, Elden Ring, Mount \& Blade, and Minecraft (melee and ranged combat). Hit SR requires at least one valid strike on a dynamic target; Kill SR requires defeating the designated target. Table~\ref{tab:cross} reports Hit SR. These metrics assess short-horizon combat rather than general game completion.

\paragraph{Metrics.}
We employ three primary metrics to assess performance and efficiency:
\textit{Success Rate (SR):} A task is considered successful if the goal is achieved.
\textit{Steps:} Average environmental steps per task completion, serving as a proxy for execution efficiency.
\textit{Latency:} The mean input-to-action time (s) per step.

\begin{table*}[t]
\centering
\small
\caption{\textbf{Evaluation on Minecraft.} Tasks are categorized into ``Embodied'', ``Combat'', and ``GUI''. We report the average steps (``Steps'') and the success rate (``SR'', \%). 
3A-VLA (full model): IA (Qwen2-VL-7B, zero-shot), ESA (SegFormer-B2, zero-shot), AAR. 
'-' denotes task failure. Bold indicates best performance.}
\label{tab:mc}
\resizebox{\textwidth}{!}{
\renewcommand\arraystretch{1.25}

\begin{tabular}{@{}lc ccc c ccc c ccc@{}}
\toprule
 & & \multicolumn{3}{c}{\includegraphics[height=1.2em]{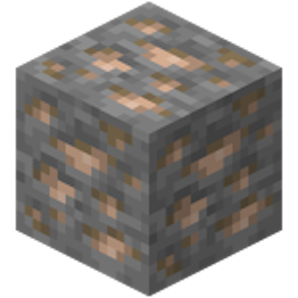} \textbf{Embodied Tasks}} & & \multicolumn{3}{c}{\includegraphics[height=1.2em]{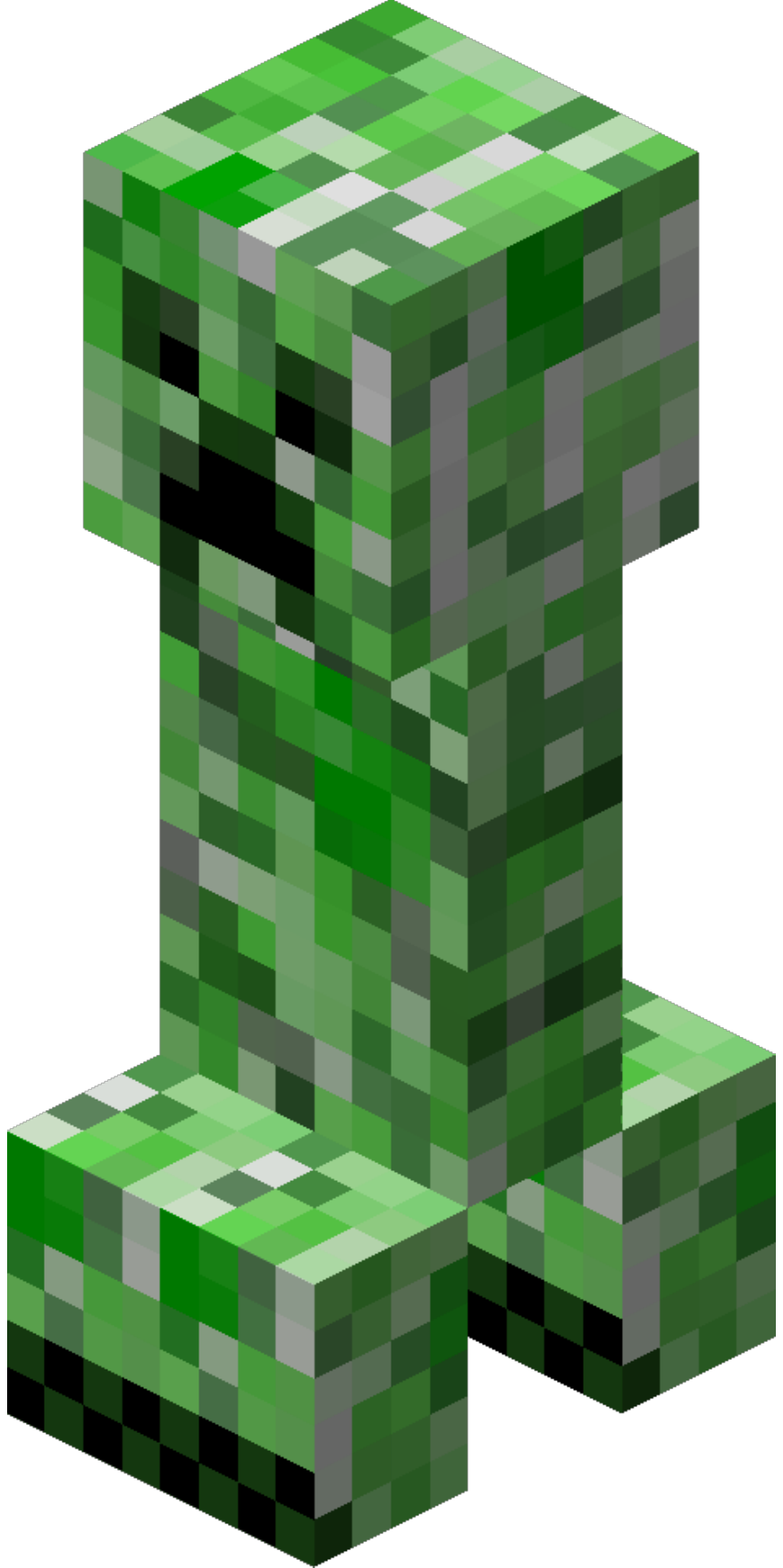} \textbf{Combat Tasks}} & & \multicolumn{3}{c}{\includegraphics[height=1.2em]{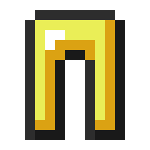} \textbf{GUI Tasks}} \\
 \cmidrule(lr){3-5} \cmidrule(lr){7-9} \cmidrule(l){11-13}
\multirow{-2}{*}{\textbf{Model}} & \textbf{Size} & Steps $\downarrow$ & SR (Mini) $\uparrow$ & SR (All) $\uparrow$ & & Steps $\downarrow$ & SR (Mini) $\uparrow$ & SR (All) $\uparrow$ & & Steps $\downarrow$ & SR (Mini) $\uparrow$ & SR (All) $\uparrow$ \\ \midrule

\multicolumn{13}{l}{\cellcolor[HTML]{ECF4FF}{ \textit{\textbf{Previous Methods}}}} \\
VPT~\cite{vpt} & 248M & 377 & 10.1$^{\pm2.8}$ & 6.0$^{\pm1.4}$ & & 396 & 3.6$^{\pm1.5}$ & 3.6$^{\pm1.2}$ & & 398 & 0.7$^{\pm0.1}$ & 0.8$^{\pm0.3}$ \\
STEVE-1~\cite{lifshitz2023steve} & 248M & 384 & 8.4$^{\pm2.4}$ & 8.0$^{\pm1.8}$ & & 395 & 4.9$^{\pm1.4}$ & 3.9$^{\pm1.1}$ & & 391 & 0.0 & 3.2$^{\pm1.3}$ \\
ROCKET-1~\cite{cai2024rocket1} & 72B & 392 & 19.2$^{\pm3.5}$ & 18.9$^{\pm3.6}$ & & 320 & 29.8$^{\pm4.2}$ & 27.9$^{\pm4.8}$ & & - & - & - \\

\midrule
\multicolumn{13}{l}{\cellcolor[HTML]{FEF1F1}{ \textit{\textbf{Vision-Language-Action Models}}}} \\
JARVIS-VLA~\cite{li2025jarvis}  & 7B & 305 & 31.0$^{\pm4.9}$ & 30.0$^{\pm5.3}$ & & 352 & 18.3$^{\pm3.1}$ & 18.5$^{\pm3.6}$ & & 339 & 25.3$^{\pm4.2}$ & 25.1$^{\pm4.5}$ \\
OpenHA~\cite{wang2025openha}
& 7B & 287 & 37.0$^{\pm5.8}$ & 30.1$^{\pm4.9}$ &
& 316 & 40.0$^{\pm6.1}$ & 31.9$^{\pm5.2}$ &
& 314 & 33.3$^{\pm5.0}$ & 32.5$^{\pm5.3}$ \\

\textbf{3A-VLA (Ours)}
& 7B & \textbf{278} & \textbf{40.6$^{\pm3.2}$} & \textbf{33.5$^{\pm2.6}$} &
& \textbf{271} & \textbf{45.9$^{\pm3.5}$} & \textbf{35.4$^{\pm2.9}$} &
& \textbf{307} & \textbf{34.7$^{\pm2.7}$} & \textbf{34.8$^{\pm2.6}$} \\
\bottomrule
\end{tabular}
}
\end{table*}

\paragraph{Training and evaluation settings.}
We fine-tune Qwen2-VL-7B~\cite{qwen2vl} independently for each environment.
For Minecraft, the training set consists of approximately 50 hours of gameplay data, whereas the Game for Peace model is trained on a separate 200-hour corpus.
In each dataset, the task instructions, our proposed IA and ESA data, and ground-truth frame-action pairs are aligned.
Unless otherwise noted, the primary benchmark results use the \emph{simplest} instantiation: IA keywords were generated in a zero-shot manner by Qwen2-VL-7B, and ESA semantic maps were produced by the pretrained SegFormer-B2~\cite{xie2021segformer} without any domain-specific fine-tuning. 
Section~\ref{sec:cost} compares the annotation costs and performance of enhanced annotators. Section~\ref{sec:pruning} evaluates pruning in a separate deployment study.
Training is distributed across 8 NVIDIA H20 GPUs, while inference is performed on a single L40S GPU.
The agent operates strictly on first-person RGB frames (640 $\times$ 360) and maps outputs to a discretized human-like action space. For ESA, we set the grid resolution to $H_g\!\times\!W_g = 8 \times 8$ with $K=2$.

\paragraph{Comparison methods.}
Minecraft baselines are VPT~\cite{vpt}, STEVE-1~\cite{lifshitz2023steve}, ROCKET-1~\cite{cai2024rocket1}, JARVIS-VLA~\cite{li2025jarvis}, and OpenHA~\cite{wang2025openha}.
For Game for Peace, we compare proprietary foundation models using zero-shot CoT, Qwen2-VL-7B, controlled Qwen2-VL-7B IL variants trained on identical data, and the CombatVLA reproduction and adapted Open-P2P listed in Table~\ref{tab:gp}.
Controlled variants use CoT and caption supervision as alternatives to IA and ESA, respectively.

\paragraph{Evaluation protocol.}
The Game for Peace comparisons among 3A-VLA, Vanilla IL, IL+CoT+Caption, CombatVLA, and Open-P2P use 100 evaluation runs with shared episode manifests, so the models face the same episodes. We reproduce CombatVLA with a 7B model and adapt the public Open-P2P checkpoint using the same gameplay data, training budget, and interface. The controlled IL variants share the Qwen2-VL-7B backbone and training data with 3A-VLA, allowing a direct comparison of supervision targets. For Minecraft ablations, we macro-average success rates over the Embodied, Combat, and GUI groups of MCU Mini.

\subsection{Main results: Analysis of agent behaviors}
\label{sec:main_results}

\paragraph{Evaluation on Minecraft.}
Table~\ref{tab:mc} shows the results for the three Minecraft task groups. 3A-VLA achieves the best overall performance: it attains the highest OOD success rates across all three task groups while requiring the fewest steps. The improvement is most pronounced in ``Combat'', where 3A-VLA increases OpenHA's success rate from 40.0\% / 31.9\% to 45.9\% / 35.4\% on ``Mini/All'' splits---demonstrating the value of abstraction under rapid, cluttered observations.
The gains extend beyond Combat to Embodied tasks, where the Mini/All SR increases from OpenHA's 37.0/30.1\% to 40.6/33.5\%, and GUI tasks, where it rises from 33.3/32.5\% to 34.7/34.8\%.
In Combat, the average number of steps decreases from 316 to 271, indicating more efficient execution. 
The relatively smaller gains on GUI suggest that abstraction is particularly beneficial for visually dynamic control tasks.

\begin{table*}[t]
\centering
\small
\caption{\textbf{Evaluation on Game for Peace.}
We compare success rate (SR, \%) and inference latency (Lat., s).
3A-VLA uses zero-shot Qwen2-VL-7B IA, pretrained SegFormer-B2 ESA, and AAR.
The controlled IL variants use Qwen2-VL-7B.
Avg. is the mean success rate across the six tasks.
CombatVLA is our reproduction and Open-P2P is adapted.}
\label{tab:gp}
\resizebox{\textwidth}{!}{
\renewcommand\arraystretch{1.2}
\setlength{\tabcolsep}{3.8pt}
\begin{tabular}{@{}lcccccccc@{}}
\toprule
\textbf{Model} &
\includegraphics[height=1.1em]{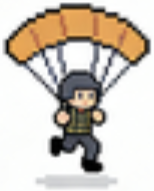} \textbf{Parachuting} &
\includegraphics[height=1.1em]{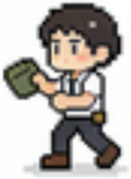} \textbf{Scavenging} &
\includegraphics[height=1.1em]{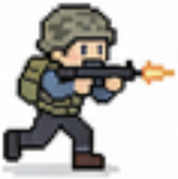} \textbf{Combat} &
\includegraphics[height=1.1em]{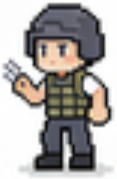} \textbf{Revival} &
\includegraphics[height=1.1em]{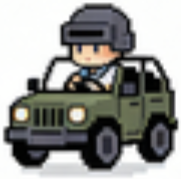} \textbf{Vehicle} &
\includegraphics[height=1.1em]{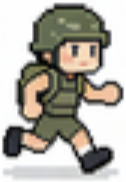} \textbf{Rotation} &
\textbf{Avg.} $\uparrow$ &
\textbf{Lat.} $\downarrow$ \\
\midrule

\multicolumn{9}{l}{
\cellcolor[HTML]{F5F9FF}
\textit{\textbf{Proprietary Foundation Models}}} \\
GPT-4o~\cite{openai2024gpt4o}
& 41.4 & 31.4 & 18.6 & 47.1 & 38.6 & 40.0 & 36.2 & 1.6 \\
Claude-3.7-Sonnet~\cite{anthropic2025claude3.7}
& 48.6 & 38.6 & 24.3 & 57.1 & 41.4 & 45.7 & 42.6 & 2.2 \\
Gemini-2.5-Pro~\cite{google2025gemini2.5}
& 55.7 & 42.8 & 25.7 & 58.6 & 47.1 & 51.4 & 46.9 & 1.5 \\

\midrule
\multicolumn{9}{l}{
\cellcolor[HTML]{F8FFF8}
\textit{\textbf{Open-Source Vision-Language Models}}} \\
Qwen2-VL-7B~\cite{qwen2vl}
& 22.9 & 15.7 & 8.6 & 27.1 & 18.6 & 21.4 & 19.1 & 1.1 \\

\midrule
\multicolumn{9}{l}{
\cellcolor[HTML]{FFF5F5}
\textit{\textbf{Vision-Language-Action Models}}} \\

Vanilla IL
& 54.3
& 48.6
& 44.3
& 68.6
& 51.4
& 52.9
& 53.4 & 1.1 \\

IL + CoT
& 55.3 & 49.1 & 45.7 & 68.1 & 53.1 & 53.9 & 54.2 & 1.1 \\

IL + Image Caption
& 56.0 & 50.4 & 46.3 & 68.7 & 53.4 & 54.3 & 54.9 & 1.1 \\

IL + CoT + Caption
& 56.3
& 50.9
& 46.8
& 69.3
& 53.7
& 55.0
& 55.3 & 1.1 \\

CombatVLA~\cite{chen2025combatvla}
& 48.1
& 44.7
& 42.1
& 44.0
& 47.1
& 49.0
& 45.8 & -- \\

Open-P2P~\cite{openp2p}
& 48.7
& 45.2
& 63.5
& 54.6
& 58.3
& 60.1
& 55.1 & -- \\

\textbf{3A-VLA (Ours)}
& \textbf{66.1}
& \textbf{60.5}
& \textbf{67.1}
& \textbf{70.2}
& \textbf{68.9}
& \textbf{65.7}
& \textbf{66.4}
& \textbf{1.1} \\

\bottomrule
\end{tabular}
}
\end{table*}

\paragraph{Evaluation on Game for Peace.}
As shown in Table~\ref{tab:gp}, 3A-VLA achieves state-of-the-art performance on the Game for Peace benchmark. The complete IA+ESA+AAR model reaches 66.4\% versus 55.3\% for IL+CoT+Caption, a gain of 11.1 percentage points, with particularly large gains in combat and vehicle acquisition. These comparisons support the benefit of the combined abstraction and reweighting pipeline.
Compared with Vanilla IL, 3A-VLA improves the average SR by 13.0 percentage points (53.4\% to 66.4\%).
Against the stronger IL+CoT+Caption baseline, the largest gains occur in Combat (46.8\% to 67.1\%, +20.3 points) and Vehicle acquisition (53.7\% to 68.9\%, +15.2 points), whereas Revival improves only modestly (69.3\% to 70.2\%, +0.9 points).
This non-uniform improvement is consistent with greater benefits in visually dynamic, action-sensitive tasks.

\subsection{Mechanism analysis}

\paragraph{Individual and combined contributions.}
On Minecraft, IA alone improves SR from 24.6\% to 30.2\%, while ESA alone reaches 30.6\%. Combining IA and ESA further raises SR to 36.5\% without AAR, and the full model achieves 40.4\%.
These results demonstrate the complementary benefits of the two abstractions, with AAR providing an additional improvement beyond their joint supervision.

\paragraph{Analysis of representation sparsity in IA and ESA.}
Table~\ref{tab:ablation} validates our ``less is more'' hypothesis by comparing sparse abstractions against their denser counterparts within an identical pipeline.
\textbf{Intention Abstraction (IA):} Contrary to the intuition that more reasoning helps, keyword-based IA surpasses CoT variants. This suggests that concise keyword semantics effectively capture reusable intent primitives, circumventing the performance degradation associated with redundant text related to the rationale.
\textbf{Environment Abstraction (ESA):} Our latent ESA outperforms dense alternatives---including action-conditioned image prediction (``Img. Pred.'') and global semantic segmentation (``Full Sem.'')---as well as non-pixel-level scene descriptions like image captions (``Img. Caption''). While captions provide spatial cues, they are often encumbered by redundant linguistic details. In contrast, ESA operates at the pixel level, filtering transient noise to prioritize structurally actionable information through coarse affordance grids, rather than pursuing dense reconstruction fidelity.
With AAR off and backbone, trajectories, semantic scope, and a 100-token target budget matched, IA+ESA reaches $36.5/62.9\%$ MC/GP SR versus $32.4/55.0\%$ for a unified abstraction, and also achieves higher means than the two alternative splits (entity/affordance and where/how).

\begin{table}[t]
\centering
\small
\begin{minipage}[t]{0.58\linewidth}
\centering
\caption{\textbf{Ablation on IA/ESA} and alternatives. Sparse keyword IA and latent ESA constitute our proposed abstractions; the bottom row shows our full approach.}
\label{tab:ablation}
\vspace{0.3em}
\renewcommand\arraystretch{1.12}
\setlength{\tabcolsep}{4pt}
\begin{tabular}{@{}lccccc@{}}
\toprule
\multirow{2}{*}{\textbf{Variant}} & \multicolumn{2}{c}{\textbf{Alternatives}} & \multicolumn{2}{c}{\textbf{Minecraft}} \\
\cmidrule(lr){2-3} \cmidrule(l){4-5}
& IA & ESA & Steps $\downarrow$ & SR $\uparrow$ \\
\midrule
\multirow{2}{*}{Baseline}
& $-$ & $-$ & 332 & 24.6 \\
& Full CoT & Img. Caption & 309 & 32.3 \\
\midrule
\multirow{2}{*}{IA}
 & Full CoT & $-$ & 310 & 28.3 \\
 & \textbf{Keywords} & $-$ & 298 & 30.2 \\
\midrule
\multirow{4}{*}{ESA}
 & $-$ & Img. Caption & 320 & 27.0 \\
 & $-$ & Img. Pred. & 322 & 26.8 \\
 & $-$ & Full Sem. & 318 & 27.4 \\
 & $-$ & \textbf{Latent} & 304 & 30.6 \\
\midrule
Ours (w/o AAR) & \textbf{Keywords} & \textbf{Latent} & 297 & 36.5 \\
\rowcolor[HTML]{D9D9D9}
\textbf{Ours} & \textbf{Keywords} & \textbf{Latent} & \textbf{285} & \textbf{40.4} \\
\bottomrule
\end{tabular}
\end{minipage}%
\hfill
\begin{minipage}[t]{0.40\linewidth}
\centering
\captionof{table}{\textbf{Ablation on AAR} and alternatives.
We compare \textit{manual} and \textit{learned} weighting schemes.
}
\label{tab:ablation2}
\vspace{0.35em}
\footnotesize
\renewcommand\arraystretch{1.12}
\setlength{\tabcolsep}{3pt}
\begin{tabular}{@{}lcc@{}}
\toprule
\multirow{2}{*}{\textbf{Scheme}} & \multicolumn{2}{c}{\textbf{Minecraft}} \\
\cmidrule(l){2-3}
& Steps $\downarrow$ & SR $\uparrow$ \\
\midrule
Ours (w/o AAR) & 297 & 36.5 \\
\midrule
Fixed $w{=}0.33$ & 303 & 36.9 \\
Fixed $w{=}0.67$ & 302 & 37.0 \\
Scalar cosine & 299 & 37.1 \\
Scalar $\odot$ & 295 & 37.6 \\
\midrule
CoT+Cap. w/o AAR & 309 & 32.3 \\
CoT+Cap. w/ VLM-Score & 307 & 32.8 \\
CoT+Cap. w/ AAR & 301 & 36.0 \\
\midrule
AAR w/o $\bar{H}$ & 291 & 38.2 \\
AAR w/o $\mathcal{D}$ & 289 & 37.6 \\
\rowcolor[HTML]{D9D9D9}
\textbf{AAR (Ours)} & \textbf{285} & \textbf{40.4} \\
\bottomrule
\end{tabular}
\end{minipage}

\end{table}

\begin{table}[t]
\centering
\small
\caption{\textbf{Zero-shot cross-game combat transfer}
(Hit SR, \%).
All models are trained on Game for Peace combat data via imitation
learning and evaluated without target-game fine-tuning.
3A-VLA uses the simplest IA/ESA pipeline.}
\label{tab:cross}
\vspace{0.3em}
\renewcommand\arraystretch{1.15}
\setlength{\tabcolsep}{4pt}
\begin{adjustbox}{max width=\linewidth}
\begin{tabular}{@{}lcccccc@{}}
\toprule
\textbf{Method}
& \textbf{Valorant}
& \textbf{Minecraft}
& \textbf{CS2}
& \textbf{GTA V}
& \textbf{Elden Ring}
& \textbf{Mount \& Blade} \\
\midrule
Qwen2-VL-7B (Vanilla IL)
& 27.1 & 8.6 & 18.6 & 15.7 & 7.1 & 8.6 \\
Qwen2-VL-7B (IL + CoT + Caption)
& 35.1 & 17.7 & 28.3 & 26.0 & 13.2 & 16.1 \\
\rowcolor[HTML]{D9D9D9}
\textbf{3A-VLA (Ours)}
& \textbf{42.9}
& \textbf{25.7}
& \textbf{40.7}
& \textbf{38.6}
& \textbf{21.4}
& \textbf{24.3} \\
\bottomrule
\end{tabular}
\end{adjustbox}
\end{table}

\paragraph{Adaptive loss reweighting via AAR.}

Table~\ref{tab:ablation2} ablates AAR within the IA-ESA pipeline.
\textbf{(i)} The ``Fixed'' rows apply a discrete threshold: if the semantic keywords from IA and ESA are non-overlapping, we manually down-weight the action loss by a static factor ($w{=}0.33$ or $0.67$) rather than a learned weight.
Scalar cosine and $\odot$ replace this discrete rule with simple heuristic alignment scores (0$\sim$1). 
These manual designs recover part of the gain, but all remain below learned AAR, showing that IA--ESA alignment should modulate action imitation in a continuous and adaptive manner. 

\textbf{(ii)} AAR remains effective even with degraded IA/ESA surrogates like ``CoT+Cap,'' consistently outperforming naive weighting strategies that rely on an additional VLM for direct action loss scoring (``VLM-Score'').
\textbf{(iii)} Removing the entropy regularizer $\bar{H}$ or stop-gradient $\mathcal{D}$ degrades performance, as AAR works best when the weight is learned from the relation feature without collapsing onto a few samples or directly perturbing the backbone representation. The integration of the IA, ESA, and learned AAR achieves the strongest performance, underscoring that sparse abstraction and adaptive reweighting are mutually reinforcing.
With IA+ESA fixed, AAR reaches $40.4\pm1.5\%$ MC SR versus $37.5\pm1.7\%$ for the strongest loss-only alternative. Its score is a learned relation-based weighting signal, not calibrated demonstration reliability.

\begin{figure}[t]
\centering
\includegraphics[width=1.0\linewidth]{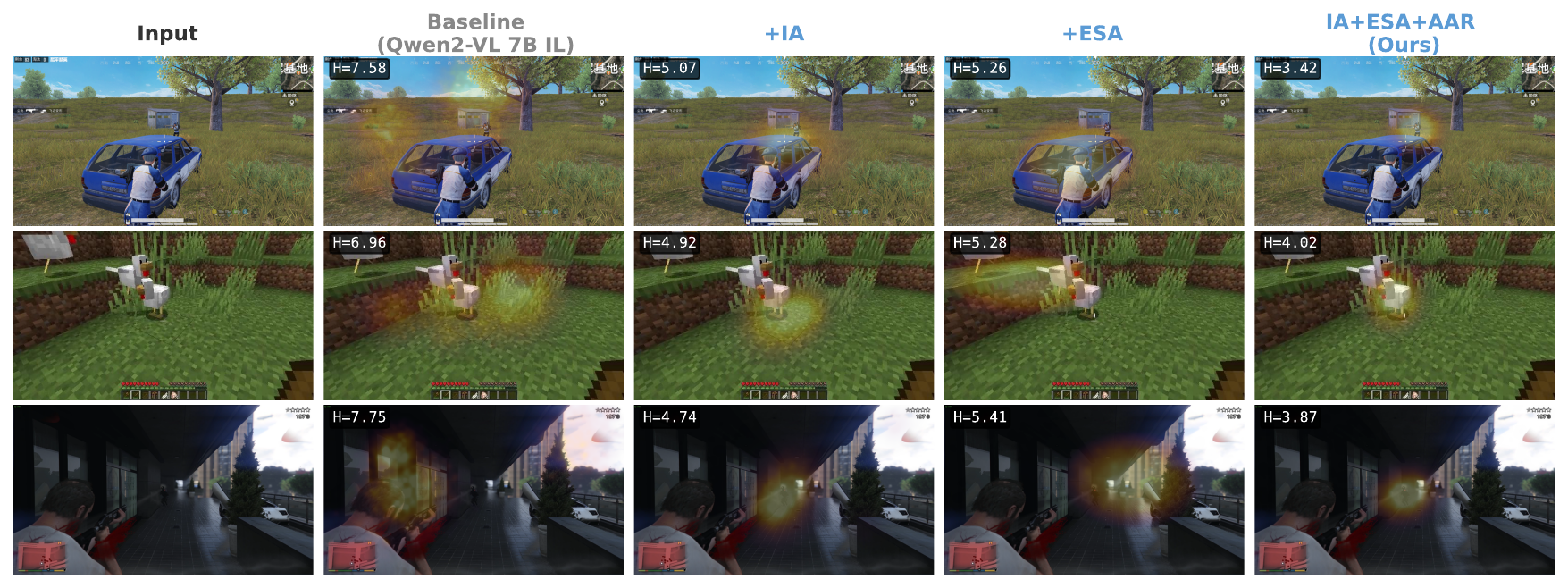}
\caption{
\textbf{Attention and zero-shot generalization.} Trained on Game for Peace, the IA, ESA, and AAR progressively sharpen the focus on task-relevant regions as the attention entropy decreases, enabling the model to suppress irrelevant visual distractors even in unseen games (rows 2 and 3).
}
\label{fig:attention}
\vspace{-6pt}
\end{figure}

\paragraph{Emergent attention sparsity as a side effect.}
Figure~\ref{fig:attention} shows more concentrated attention with abstraction supervision: entropy decreases from $H=7.58$ to $H=3.42$ in the first example. Section~\ref{sec:pruning} evaluates whether this concentration supports token pruning, comparing connectivity scoring with generic pruning methods and reporting the accuracy--latency trade-off.

\subsection{Zero-shot combat transfer and robustness}
Table~\ref{tab:cross} reports Hit SR for GP-combat-trained models without target-game adaptation. 3A-VLA attains the highest mean Hit SR in all six games, e.g., 42.9\% versus 39.4\% for adapted Open-P2P in Valorant, with the largest relative gains over Vanilla IL in visually distant titles such as Minecraft (8.6\% to 25.7\%). 3A-VLA's Kill SR ranges from $10.6\%$ to $21.4\%$ across the six games.

In Minecraft controls that pair opposing instructions in the same scene, 3A-VLA successfully executes both instructions in 7.5/14.3/17.8\% of wall/doorway/enemy trials, compared with 2.3/4.2/5.1\% for random instructions. This supports partial instruction-conditioned disambiguation, with a remaining gap to the Minecraft-trained reference. The evidence supports combat transfer under this protocol. Target-title exposure during pretraining cannot be excluded. Occlusion, long-horizon recovery, and domain-specific interactions remain challenging.

\subsection{Component cost-effectiveness analysis}
\label{sec:cost}

Table~\ref{tab:cost_detail} shows that the simplest IA+ESA+AAR pipeline achieves 40.4\% SR with approximately 8 GPU-hours of annotation, improving over the baseline by 15.8 percentage points (about 5 GPU-hours for IA and 3 for ESA, with no API cost or manual labeling).
It also exceeds either enhanced component alone (34.8\% for IA; 32.5\% for ESA). 
Enhancing both annotators further raises SR to 45.0\% (+4.6 points).

With Qwen3-VL-8B, 3A-VLA also improves over the corresponding imitation-learning baseline on both Minecraft (+16.2 points) and Game for Peace (+10.5 points), supporting applicability across backbones and environments.

\subsection{Inference efficiency and token pruning}
\label{sec:pruning}

\paragraph{Output order and latency.}
We vary the output order while keeping the rest of the model configuration fixed and measure end-to-end input-to-action latency on one L40S over 100 evaluation runs. With actions first, inference stops after the action chunk and achieves 40.4\% success on Minecraft at 1.1\,s per step. Decoding IA before actions raises latency to 1.3\,s with 39.9\% success. Decoding both IA and ESA before actions takes 3.2\,s and yields 37.6\% success. This comparison motivates action-first decoding for deployment.

\begin{figure}[t]
\centering
\normalsize
\setlength{\parskip}{0pt}
\captionsetup{
  font=normalsize,
  justification=raggedright,
  singlelinecheck=false,
  skip=4pt
}

\begin{minipage}[t]{0.55\textwidth}
  \vspace{0pt}
  \centering
  \includegraphics[width=\linewidth]{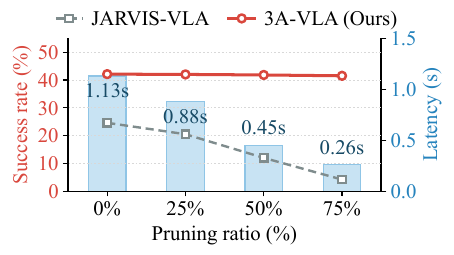}
  \caption{\textbf{Robustness to token pruning.}
  3A-VLA maintains success rate at 75\% pruning
  with a $4\times$ inference speedup.}
  \label{fig:prune}
\end{minipage}\hfill%
\begin{minipage}[t]{0.4\textwidth}
  \vspace{0pt}
  \centering
  \normalsize
  \captionsetup{position=top}
  \captionof{table}{\textbf{Pruning on Minecraft.}
  75\% pruning with enhanced IA+ESA;
  Connectivity scoring achieves the best success rate among pruning methods.}
  \label{tab:prune_cmp}

  \setlength{\tabcolsep}{7.5pt}
  \renewcommand{\arraystretch}{1.05}
  \begin{tabular}{@{}lcc@{}}
    \toprule
    Method & SR (\%) & Latency (s) \\
    \midrule
    No pruning & 42.1 & 1.1 \\
    \rowcolor{gray!10}
    \textbf{Connectivity} & \textbf{41.5} & \textbf{0.3} \\
    FastV~\cite{chen2024fastv} & 39.2 & 0.3 \\
    ToMe~\cite{bolya2022token} & 38.6 & 0.3 \\
    Magnitude & 36.4 & 0.3 \\
    Random & 30.9 & 0.3 \\
    \bottomrule
  \end{tabular}
\end{minipage}

\end{figure}

\paragraph{Visual token pruning.}
Figure~\ref{fig:prune} illustrates the robustness of the enhanced IA+ESA configuration (without AAR).
While JARVIS-VLA degrades substantially under increasing pruning ratios, our abstraction-trained representation retains most of its performance.
At 75\% pruning, connectivity-based selection
achieves 41.5\% SR compared with 42.1\% without
pruning, with approximately 4$\times$ faster inference.

Table~\ref{tab:prune_cmp} further compares connectivity scoring with FastV and ToMe under the same model configuration and latency budget.
The baseline degrades under the tested pruning ratios.
At equal pruning ratios, FastV and ToMe reach 39.2\% and 38.6\% SR, compared with 41.5\% for connectivity scoring.
Our contribution is therefore not a new pruning algorithm, but an abstraction training recipe that makes the representation structurally prunable---the connectivity score simply exploits that structure.

\begin{table}[t]
    \centering
    \caption{\textbf{Cost-effectiveness analysis on Minecraft.} Each row reports a single configuration; the cost column shows incremental annotation effort beyond the trajectory dataset.}
    \label{tab:cost_detail}
    \vspace{0.5em}
    \begin{tabular}{llcc}
        \toprule
        \textbf{Tier} & \textbf{Configuration} & \textbf{MC SR (\%)} & \textbf{Cost} \\
        \midrule
        Baseline & No IA, no ESA, no AAR & 24.6 & --- \\
        \midrule
        Simplest & IA only (Qwen2-VL-7B, no RAG) & 30.2 & $\approx$5 GPU-hrs \\
        Simplest & ESA only (zero-shot SegFormer-B2) & 30.6 & $\approx$3 GPU-hrs \\
        Simplest & IA + ESA & 36.5 & $\approx$8 GPU-hrs \\
        \rowcolor{gray!10}
        \textbf{Simplest (Ours)} & \textbf{IA + ESA + AAR} & \textbf{40.4} & $\approx$8 GPU-hrs \\
        \midrule
        Enhanced & IA only (GPT-4o + full RAG) & 34.8 & $\approx$\$260 \\
        Enhanced & ESA only (SegFormer-B2 finetuned) & 32.5 & $\approx$16 person-hrs \\
        Enhanced & IA + ESA  & 42.1 & $\approx$\$260 + 16 p-hrs \\
        \rowcolor{gray!10}
        \textbf{Enhanced} &  \textbf{IA + ESA + AAR} &  \textbf{45.0} & $\approx$\$260 + 16 p-hrs \\
        \bottomrule
    \end{tabular}
    \vspace{0.2em} 
\end{table}

\section{Conclusion}
3A-VLA uses sparse intention and environment abstractions to guide action learning in dynamic games. AAR weights the imitation loss by the alignment between these abstractions. Experiments on Minecraft and Game for Peace show gains over the evaluated baselines, and cross-game evaluations demonstrate zero-shot combat transfer. Ablations support the contributions of sparse abstraction and adaptive weighting. Both abstractions serve as auxiliary training targets, preserving the standard VLA inference pipeline.

\textbf{Limitations and future work.}
3A-VLA remains single-frame and reactive, lacking explicit temporal memory and long-horizon planning, which limits autonomy in extended FPS tasks.
Moreover, ESA lacks explicit terrain and collision geometry, limiting fine-grained traversability reasoning.
In future work, the integration of intrinsic motivation will be explored to extend 3A-VLA toward an autonomous strategic discovery model.

\clearpage
\section*{Acknowledgments and Disclosure of Funding}

This work was supported by Tencent through the provision of computational resources, data infrastructure, and research support. Additional funding was provided by the National Natural Science Foundation of China (Grant No.~52375271) and the ``Pioneer and Leading Goose'' R\&D Program of Zhejiang Province (Grant No.~2024C01044).

\noindent\textbf{Competing interests.}
The authors declare no competing interests beyond
the affiliations and support disclosed above.

\bibliographystyle{plainnat}
\bibliography{main}

\end{document}